\pdfoutput=1
\pdftrailerid{}
\documentclass{article}
\usepackage{iclr2027_conference,times}
\usepackage[utf8]{inputenc} %
\usepackage[T1]{fontenc}    %
\usepackage{hyperref}       %
\usepackage{url}            %
\usepackage{graphicx}       %
\usepackage{booktabs}       %
\usepackage{amsfonts}       %
\usepackage{nicefrac}       %
\usepackage{microtype}      %
\usepackage[table]{xcolor}  %
\usepackage[most]{tcolorbox}
\usepackage{amsmath}
\usepackage{multirow}
\usepackage{makecell}
\usepackage{placeins}
\usepackage{float}         %
\usepackage{wrapfig}       %
\usepackage{capt-of}       %
\usepackage{needspace}     %

\newcommand{\modelfullname}{Autoregressive Thought Flow}
\newcommand{\model}{ATF}

\title{Next Thoughts Are Distributions: Generative Autoregressive Reasoning in the Latent Space}

\author{%
  \normalfont Yang LI$^{1}$, Yi Wang$^{1}$, Shiyuan Huang$^{2}$, Yang Liu$^{2}$, Hao Wang$^{3}$, Chengzhi Mao$^{1}$\\[2pt]
  \normalfont $^{1}$Rutgers University\\
  \normalfont $^{2}$Amazon\\
  \normalfont $^{3}$University of Illinois at Urbana-Champaign
}

\AtBeginDocument{%
  \setlength{\abovedisplayskip}{5pt plus 2pt minus 2pt}%
  \setlength{\belowdisplayskip}{5pt plus 2pt minus 2pt}%
}
\usepackage{etoolbox}
\iclrfinalcopy
\makeatletter
\patchcmd{\@maketitle}
  {Published as a conference paper at ICLR 2027}
  {Preprint}
  {}
  {\PackageError{arxiv-preprint}{Unable to set preprint header}{Check the title macro before publishing.}}
\makeatother
\hypersetup{
  pdftitle={Next Thoughts Are Distributions: Generative Autoregressive Reasoning in the Latent Space},
  pdfauthor={Yang LI, Yi Wang, Shiyuan Huang, Yang Liu, Hao Wang, Chengzhi Mao}
}

\begin{document}
\maketitle

\begin{abstract}
Reasoning problems often admit multiple valid ways to proceed.
Continuous reasoning promises to move computation beyond language tokens into a more compact latent space, but representing several plausible ways to think next remains difficult.
We introduce \modelfullname{} (\model{}), which models the next continuous thought as a multimodal distribution.
A causal autoregressive model performs the reasoning computation, while a lightweight diffusion head generates a plausible next thought from the resulting condition.
The sampled thought is fed back into the model, allowing continuous reasoning to unfold for a variable number of steps while preserving the pretrained backbone.
Across mathematical reasoning tasks, \model{} improves accuracy with compact latent traces and benefits from reinforcement learning and additional test-time thinking.
Multi-sample evaluation shows broader solution coverage, indicating that its multimodal predictions capture useful diversity among reasoning paths.
Our results suggest that continuous reasoning is more effective when multiple possible next thoughts remain available rather than being collapsed into a single prediction.

\end{abstract}

\section{Introduction}

The next thought is rarely unique. The same problem can support several valid lines of reasoning, each following a different sequence of intermediate computations. Chain-of-thought makes these computations easy to supervise and inspect, but also requires every step to be expressed through a discrete, sequential language interface~\citep{wei2022chain}.
Continuous reasoning offers a different possibility: keep intermediate computation in latent space~\citep{hao2022training}, where thoughts can be compact and need not be expressed in words.

This freedom creates a basic problem: what should come next?
In language, a softmax defines a distribution over a fixed vocabulary.
In continuous space, there is no fixed set of alternatives, and several distinct next thoughts may all be valid.
Existing methods learn such latent states in different ways~\citep{softthinking,kong2025latent,hao2022training}, including by reconstructing or predicting target representations~\citep{ray2025mulltokens,barrault2024large,tan2025think,cheng2024compressed,shao2025continuous}.
Recent work introduces stochastic transitions or explores multiple latent trajectories~\citep{zou2026latent,zhou2026lepo,wu2026llms,you-etal-2026-parallel}.
Diffusion provides a natural way to model multimodal distributions over continuous outputs~\citep{li2024autoregressive,zhou2024transfusion,team2025nextstep,deng2024causal}, but diffusion-based reasoning often replaces the causal autoregressive interface with bidirectional or full-sequence denoising~\citep{kang2025ladir}.
This changes the structure of pretrained language models and makes variable-length reasoning less natural.
The challenge is therefore to preserve this multimodality without giving up autoregressive reasoning.

\begin{figure}[t]
    \centering
    \includegraphics[width=\linewidth]{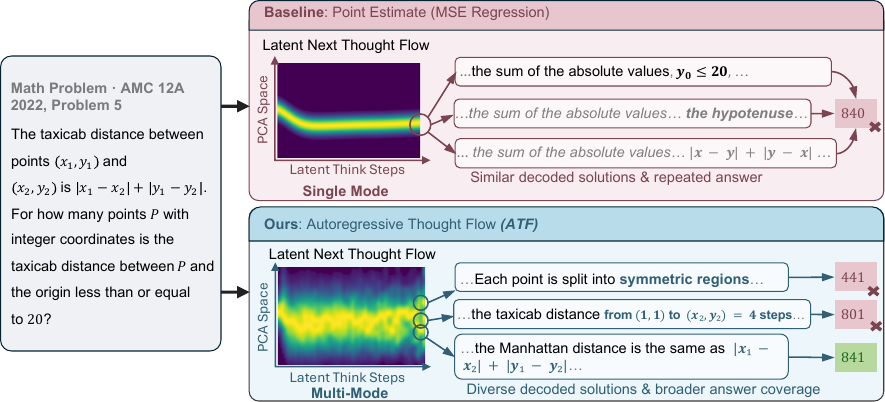}
    \caption{\textbf{Our \model{} enables multiple latent continuations.}
    Given the same math prompt, a point-estimate continuous-thought model collapses the next-thought distribution into a single averaged trajectory, producing similar decoded continuations and an incorrect final answer. \model{} instead samples next thoughts from a multi-mode latent distribution and feeds each realized thought back autoregressively, allowing distinct reasoning paths to be explored. In this example, one sampled \model{} trajectory follows the correct taxicab-counting path and reaches the correct answer.}
    \label{fig:teaser}
\end{figure}

Our key idea is to model the next thought as a multimodal distribution while keeping reasoning autoregressive.
We introduce \modelfullname{} (\model{}), where a causal transformer performs the reasoning computation and produces a condition for what should come next. 
A lightweight diffusion head~\citep{li2026back} turns this condition into a distribution over multiple plausible continuous thoughts, samples one, and feeds it back into the model. See a concrete example in Figure~\ref{fig:teaser}.
The backbone therefore determines what is plausible, while diffusion preserves the alternative ways that the next thought can be realized.
Because thoughts are generated one at a time, \model{} retains the structure of pretrained causal models and is able to extend reasoning without fixing the trajectory length in advance.

This design provides several additional benefits.
Sampling from the next-thought distribution naturally exposes multiple reasoning trajectories that can be explored at test time.
The diffusion formulation also inherits classifier-free guidance~\citep{ho2022classifier} at no architectural cost, providing a built-in control over exploration at inference time.
Finally, retaining the autoregressive interface makes it straightforward to apply reinforcement learning directly to the model's own unrolled latent trajectories.

Experiments and visualizations show that these capabilities translate into stronger reasoning.
Across mathematical reasoning benchmarks, including the challenging OlympiadBench~\citep{he2024olympiadbench} and AMC~12~\citep{maa_competitions}, \model{} improves the accuracy--reasoning-length trade-off over prior continuous-reasoning methods and short-budget language chain-of-thought~\citep{wei2022chain}, while using compact latent traces.
Reinforcement learning further improves accuracy and leads the model to use more latent computation when useful, while generating additional thoughts at test time continues to improve performance.
In multi-sample evaluation, \model{} improves Pass@100 over the strongest prior baseline by 19.3 points on average across MATH-500~\citep{hendrycks2021math}, GSM8K~\citep{cobbe2021gsm8k}, and OlympiadBench~\citep{he2024olympiadbench}, outperforming iCoT~\citep{icot}, Coconut~\citep{hao2022training}, CODI~\citep{shen2025codi}, TaH+~\citep{tahplus}, Discrete Latent Reasoning~\citep{discrete_latent_thought}, Soft Thinking~\citep{softthinking}, and the diffusion-based LaDiR~\citep{kang2025ladir}.
On OlympiadBench alone, the gain reaches 34.6 points.
Even when the final answer is decoded greedily, sampling continuous thoughts substantially increases solution coverage, showing that the diversity comes from latent reasoning itself rather than text sampling.
Decoded trajectories further reveal multiple plausible continuations from the same context.
Our results show that preserving multimodality in next-thought prediction is useful for continuous reasoning.

\section{Related Work}

\paragraph{Continuous and latent reasoning.}
Chain-of-thought exposes intermediate reasoning in language~\citep{wei2022chain}, but long textual traces can be costly at inference time.
Continuous-reasoning methods instead move this computation into latent space.
Coconut feeds hidden states back into the model~\citep{hao2022training}, Soft Thinking propagates soft token distributions~\citep{softthinking}, latent-thought models introduce variational states~\citep{kong2025latent}, and LCM~\citep{barrault2024large,duquenne2023sonar}, CCOT~\citep{cheng2024compressed}, MuLTI~\citep{ray2025mulltokens}, CoLaR~\citep{tan2025think}, CALM~\citep{shao2025continuous}, and MARCOS~\citep{liu2025marcos} learn compact continuous representations of intermediate reasoning.
These works establish continuous thoughts as a useful reasoning interface, but many still propagate a single latent continuation or optimize toward a single target representation.
Recent analyses show that such continuous reasoning can become effectively single-threaded~\citep{wu2026llms}, motivating methods that explicitly introduce stochasticity.
Latent Thought Flow uses GFlowNets to model stochastic latent trajectories~\citep{zou2026latent}, LEPO uses Gumbel-Softmax sampling to explore latent reasoning policies~\citep{zhou2026lepo}, and parallel latent reasoning injects dropout and Gaussian noise to generate multiple trajectories~\citep{you-etal-2026-parallel}.
\model{} takes a complementary approach: it directly models the next latent thought as a conditional multimodal distribution rather than introducing stochasticity around a single predicted state.

\paragraph{Generative next-state modeling.}
Diffusion and flow models provide flexible tools for modeling complex continuous distributions~\citep{song2019generative,ho2020denoising,lipman2023flow,liu2023flow}.
Recent autoregressive image and multimodal models use such generative heads for next-vector prediction without discretizing continuous outputs~\citep{li2024autoregressive,zhou2024transfusion,team2025nextstep,deng2024causal}.
LatentLM is especially related: it combines a causal Transformer with next-token diffusion to autoregressively generate continuous VAE latents~\citep{sun2024multimodal}.
Its focus is multimodal generation and understanding, whereas \model{} applies this generative interface to reasoning, where each sampled vector becomes the next intermediate thought.
LaDiR also applies diffusion to latent reasoning, but denoises a reasoning trajectory at the sequence level~\citep{kang2025ladir}.
In contrast, \model{} models the immediate next thought with a lightweight diffusion head while preserving autoregression across thoughts, allowing the reasoning trajectory to grow dynamically.
This formulation also inherits classifier-free guidance~\citep{ho2022classifier}, providing an inference-time control over latent exploration without changing the causal backbone.

\section{Method}
\label{sec:method}

Let $X$ be the input prompt, $Z=\{z_1,\ldots,z_T\}$ a sequence of continuous thoughts, and $Y$ the final answer. \model{} generates $Z$ autoregressively before decoding $Y$. For supervision, we split each text rationale into ordered chunks and use a frozen Causal-Register VAE (CR-VAE; Section~\ref{sec:crvae}) to encode them into latent registers. With $z_{\mathrm{stop}}$ denoting the embedding of the closing marker, each training example has the form
\begin{equation}
    \underbrace{\text{[Prompt]}}_{X}
    \quad\text{\texttt{<latent>}}\quad
    \underbrace{\text{[Continuous thoughts]}}_{z_1,\ldots,z_T}
    \quad\underbrace{\text{\texttt{</latent>}}}_{z_{\mathrm{stop}}}\quad
    \underbrace{\text{[Answer]}}_{Y}.
    \label{eq:hybrid-sequence}
\end{equation}
Discrete tokens use the backbone's embedding table, while continuous registers are projected into the same input-embedding space.

\subsection{Autoregressive Thought Flow}
\label{sec:continuous_reasoning}
\label{sec:latent_diffusion}

\begin{figure}[t]
    \centering
    \includegraphics[width=\linewidth]{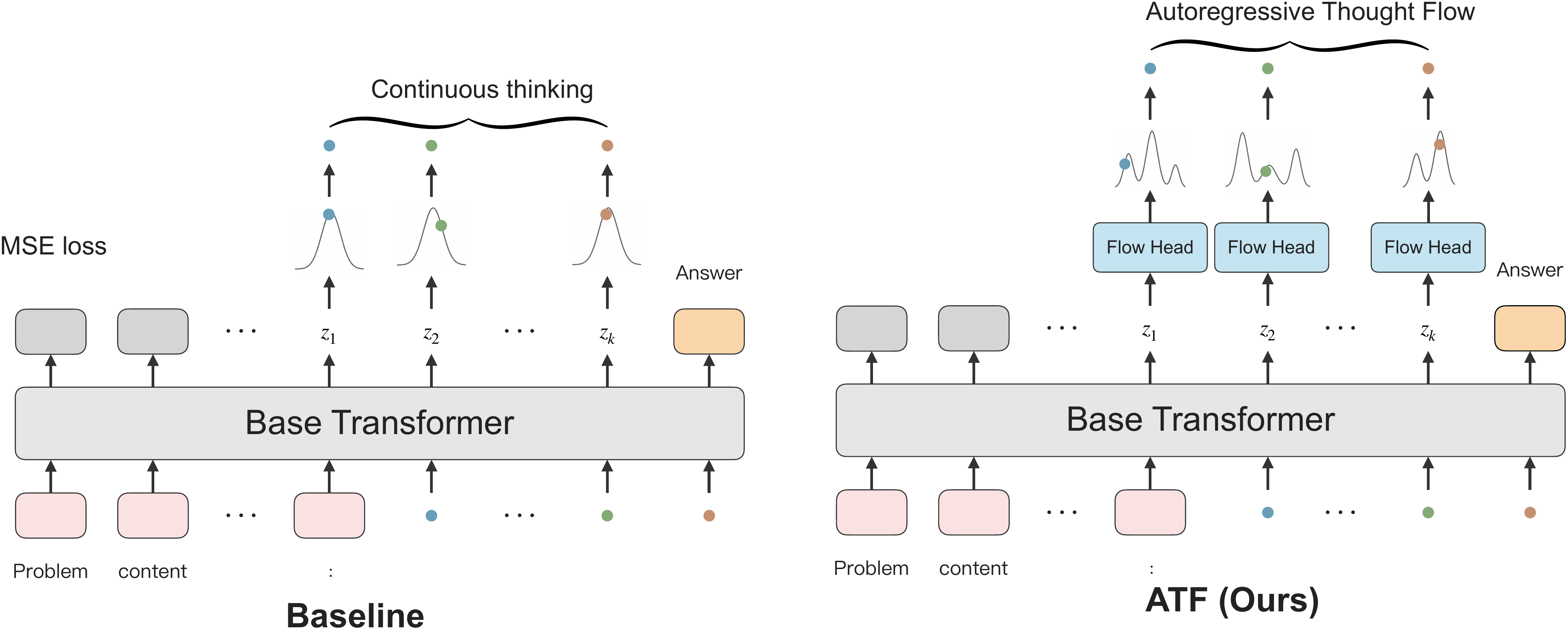}
    \caption{\textbf{\model{} generates each next thought conditioned on the realized history.}
    A causal backbone produces the condition for a lightweight flow head, which samples a continuous thought and feeds it back into the reasoning history.}
    \label{fig:method}
\end{figure}

\model{} models each next thought as a conditional multi-mode distribution, factorizing the latent trajectory as
\begin{equation}
    p_{\theta}(Z \mid X) = \prod_{t=1}^{T} p_{\theta}(z_t \mid z_{<t}, X).
\end{equation}
A causal transformer reads the prompt and previously realized thoughts to produce a context state $h_t\in\mathbb{R}^d$ at the position immediately before $z_t$. A lightweight conditional flow head~\citep{li2026back} maps this state to a distribution over the next continuous thought. Sampling a thought and feeding it back preserves autoregression across steps while allowing different continuations from the same history (Figure~\ref{fig:method}).

We train the generative head using Rectified Flow~\citep{li2026back} due to its linear trajectories and effectiveness for high-dimensional vector prediction. For a target register $z_t$ and noise $\epsilon\sim\mathcal{N}(0,I)$, the interpolation at flow time $\tau$ is
\begin{equation}
    Z_\tau = \tau z_t + (1 - \tau) \epsilon, \quad \tau \in [0, 1].
\end{equation}
The conditioned vector field $v_\psi(Z_\tau,\tau,h_t)$ predicts the noise-to-target direction through
\begin{equation}
    \mathcal{L}_{\mathrm{flow}} = \mathbb{E}_{t, \tau, \epsilon}\left[ \left\| v_\psi(Z_\tau, \tau, h_t) - (z_t - \epsilon) \right\|_2^2 \right].
\end{equation}
We evaluate multiple flow times in parallel to provide lower-variance training signals to the conditioning state $h_t$.

On the hybrid sequence in Eq.~\eqref{eq:hybrid-sequence}, we jointly train continuous-thought prediction and discrete-token generation:
\begin{equation}
    \mathcal{L}_{\mathrm{total}} = \mathcal{L}_{\mathrm{CE}}(X_{\mathrm{discrete}}) + \lambda \mathcal{L}_{\mathrm{flow}}(Z_{\mathrm{latent}}).
\end{equation}
We also supervise termination from the causal state $h_t$, so the model can decide when to end latent reasoning without reconstructing its thoughts into text. The closing marker supplies the boundary embedding $z_{\mathrm{stop}}$ for answer generation; the rollout and stopping procedure are described in Section~\ref{sec:generation}.

\subsection{Reinforcement Learning for Autoregressive Thought Flow}
\label{sec:atf-rl}

  The generative next-thought distribution provides multiple latent continuations
that can be explored under an answer reward. As an optional refinement after
SFT, we use group-relative optimization~\citep{shao2024deepseekmath} to reinforce
successful self-generated trajectories. Two levels of randomness matter: the
latent reasoning path and the text answer decoded from it. Our formulation
supports credit assignment at both levels while retaining the autoregressive flow model.

\paragraph{Rewarding latent exploration.}
For each prompt, we sample $N$ latent trajectories and decode $M$ text answers
from each trajectory. Let $r_{n,m}\in\{0,1\}$ denote answer correctness. We
score a latent history by its mean answer reward and form separate advantages:
\begin{equation}
\begin{aligned}
 \bar r_n&=\frac{1}{M}\sum_m r_{n,m},\qquad
 b=\frac{1}{N}\sum_n\bar r_n,\\
 A^{\mathrm{lat}}_n&=\bar r_n-b,\qquad
 A^{\mathrm{text}}_{n,m}=r_{n,m}-\bar r_n.
\end{aligned}
\label{eq:rl-tree-credit}
\end{equation}
The implementation standardizes and clips these advantages. Averaging sibling
answers reduces the influence of one lucky text completion on latent credit;
comparing those answers within a shared history isolates the text-decoding
contribution.

\paragraph{Updating the generative policy.}
The text-policy objective rewards correct answers conditioned on self-generated
latent histories, training the backbone to better use these histories.
Optionally, a latent-policy objective assigns outcome-based credit directly to
the autoregressive thought flow, favoring continuations that lead to successful
answers. For this latent term, we follow Flow-GRPO~\citep{liu2025flowgrpo} and
inject noise at selected flow steps to record and score stochastic transitions.
These transitions receive the latent advantage, while answer tokens receive
the text advantage. With $\mathcal C$ denoting the clipped policy-update
surrogate, we optimize
\begin{equation}
 \mathcal J_{\mathrm{RL}}=
 \lambda_z\langle\mathcal C(\rho^z,A^{\mathrm{lat}})\rangle_z+
 \lambda_y\langle\mathcal C(\rho^y,A^{\mathrm{text}})\rangle_y.
\end{equation}
Here $\rho^z$ and $\rho^y$ compare current and behavior-policy transition scores
and answer-token probabilities, respectively. Setting $\lambda_z=0$ yields
answer-policy refinement of the shared backbone; enabling the latent term
additionally updates the flow head and its conditioning backbone. The CR-VAE
remains fixed. Transition-score definitions and normalization are given in
Appendix~\ref{sec:rl-implementation}; multi-sample solution coverage assesses
the effect on exploration.

\subsection{Test-Time Generation and Guidance}
\label{sec:generation}

Given a prompt, the backbone produces $h_t$ from the realized history. We draw $\epsilon\sim\mathcal{N}(0,I)$ and generate the next thought by integrating the conditioned flow from $\tau=0$ to $\tau=1$:
\begin{equation}
    z_t = \epsilon + \int_{0}^{1} v_\psi(Z_\tau, \tau, h_t) d\tau.
\end{equation}
The sampled $z_t$ is appended to the history, updating the backbone's KV cache for the next step. Reasoning continues until the stop-margin criterion fires or the maximum latent budget is reached. The model then appends $z_{\mathrm{stop}}$ and decodes the answer through its language modeling head. This procedure applies to both the supervised model and its RL-refined variants.

\paragraph{Length and sampling budgets.}
Additional inference compute can extend a trajectory or explore more trajectories. Adjusting the stopping margin permits more latent steps before answering; Figure~\ref{fig:latent-scaling} evaluates the resulting length--accuracy trade-off for \model{} and ATF-RL. Independently sampling latent trajectories instead explores alternative continuations of the same prompt, even with greedy text decoding. The multi-sample coverage results in Table~\ref{tab:rebuttal-ladir} and Appendix~\ref{sec:multi-sample-analysis} assess whether these alternatives reach correct answers on more problems.

\paragraph{Guidance controls exploration.}
Classifier-Free Guidance (CFG)~\citep{ho2022classifier} adjusts the concentration of the next-thought distribution. During training, we drop $h_t$ with probability $p_{\mathrm{uncond}}$ to learn an unconditional field alongside the conditioned one. At inference, we use
\begin{equation}
    \tilde{v}_\tau = v_\psi(Z_\tau, \tau, \emptyset) + \gamma \left( v_\psi(Z_\tau, \tau, h_t) - v_\psi(Z_\tau, \tau, \emptyset) \right).
\end{equation}
Larger $\gamma$ amplifies the conditional component; smaller $\gamma$ retains more influence from the unconditional field. Each guided sample enters the causal history for subsequent steps. Table~\ref{tab:cfg-tradeoff} evaluates how this control trades single-sample accuracy against trajectory diversity and multi-sample coverage.

\subsection{Grounding Latent Thoughts via Causal-Register VAE}
\label{sec:crvae}

To train \model{}, we require ground-truth sequences of continuous latent thoughts $Z$ that correspond to valid reasoning steps. We extract these trajectories using an independent, pre-trained Causal-Register VAE (CR-VAE; Figure~\ref{fig:crvae-architecture} in Appendix~\ref{sec:crvae-architecture}).
Traditional block-based VAEs encode sequence chunks independently. While sufficient for isolated representations, independent encoding is fundamentally incompatible with autoregressive sequence modeling; it duplicates information across steps and breaks the causal chain. Instead, CR-VAE models the reasoning sequence as a cumulative causal process. We insert a stochastic register $z_t$ after each reasoning block, tasked with summarizing the new reasoning information available up to that point: $ q_{\phi}(z_t \mid X, x_{\le t}, z_{<t}) = \mathcal{N}\left(\mu_t, \operatorname{diag}(\sigma_t^2)\right)$.
By explicitly conditioning the latent distribution on the history of previous registers $z_{<t}$, the model is forced to encode only the \emph{residual} information introduced by the new block. The decoder then reconstructs the original sequence from this ordered stream of registers. The model is optimized using the standard evidence lower bound (ELBO):
$   \mathcal{L}_{\mathrm{CR\text{-}VAE}} = \mathcal{L}_{\mathrm{rec}} + \beta \mathcal{L}_{\mathrm{KL}}$.

Once trained, the CR-VAE is frozen. It serves as an offline tokenizer for reasoning, converting raw text demonstrations into ordered, uncertainty-aware latent trajectories that serve as the target data $Z$ for training \model{}. The CR-VAE is not used to choose final answers during benchmark evaluation; final correctness is measured from the text produced by the autoregressive decoder after latent thinking terminates. The frozen decoder instead provides an interpretability lens: we can decode any generated continuous thought $\hat{z}_t$ back into natural language to understand the internal reasoning process.

\section{Experiments}

\paragraph{Baselines.} We compare discrete CoT~\citep{wei2022chain}, matched MSE regression, and \model{} under the same Qwen3 backbones (Table~\ref{tab:main-results}). CoT uses unrestricted or capped intermediate text budgets; the zero-thought ATF ablation tests the contribution of latent thinking. MSE shares CR-VAE targets, adaptive stopping, and the final text decoder with ATF, replacing only the flow head with deterministic regression to isolate generative next-thought modeling. We directly reuse baseline scores reported in LaDiR~\citep{kang2025ladir} and CoLaR~\citep{tan2025think}, using their respective LLaMA-3.1-8B and LLaMA-3.2-1B-Instruct backbones and benchmark suites. These comparisons retain source-specific training procedures, including CoLaR's GSM8K-Aug/MATH recipe; Appendix~\ref{sec:baseline-protocols} details objectives, score sources, and protocols.

\paragraph{Datasets.} Qwen3 training mixes OpenMathReasoning~\citep{openmathreasoning} OpenR1Math~\citep{openr1math}, and DeepScaleR~\citep{deepscaler2025}. Table~\ref{tab:main-results} evaluates GSM8K~\citep{cobbe2021gsm8k} ($1{,}319$ questions), MATH-500~\citep{hendrycks2021math} ($500$), the canonical single-answer OlympiadBench~\citep{he2024olympiadbench} split ($581$), and AMC~12~\citep{maa_competitions} ($83$), spanning grade-school and competition mathematics. We additionally use AIME~24/25 to study guidance. LLaMA training uses DART-MATH-derived data for the 8B model and GSM8K-Aug/MATH for the 1B model; evaluation splits are detailed in Appendix~\ref{sec:baseline-protocols}.

\begin{table*}[t]
\centering
\hypersetup{hidelinks}
\fontsize{9}{10.5}\selectfont
\caption{\textbf{ATF improves compact reasoning over discrete and regression baselines.} Accuracy (\%) and generation lengths on Qwen3 backbones, with saved answers rescored using a common grader. Budget limits intermediate text tokens for discrete CoT and latent steps for continuous methods. $N_{\rm think}$ and $N_{\rm ans}$ report the corresponding realized thinking length and final-answer tokens. ATF-RL applies answer-policy refinement. Bold and underline mark best and second-best accuracy within each model size, excluding unbounded discrete CoT.}
\label{tab:main-results}
\setlength{\tabcolsep}{1.8pt}
\renewcommand{\arraystretch}{1.08}
\begin{tabular}{@{}ll*{12}{r}@{}}
\toprule
\textbf{Method} & \textbf{Budget} & \multicolumn{3}{c}{\textbf{GSM8K}} & \multicolumn{3}{c}{\textbf{MATH-500}} & \multicolumn{3}{c}{\textbf{OlympiadBench}} & \multicolumn{3}{c}{\textbf{AMC 12}} \\
\cmidrule(lr){3-5}\cmidrule(lr){6-8}\cmidrule(lr){9-11}\cmidrule(lr){12-14}
& & \textbf{Acc.} & $N_{\rm think}$ & $N_{\rm ans}$ & \textbf{Acc.} & $N_{\rm think}$ & $N_{\rm ans}$ & \textbf{Acc.} & $N_{\rm think}$ & $N_{\rm ans}$ & \textbf{Acc.} & $N_{\rm think}$ & $N_{\rm ans}$ \\
\midrule
\multicolumn{14}{l}{\textbf{\textit{Qwen3-1.7B}}} \\
\addlinespace[1pt]
\textit{Discrete CoT} & $\infty$ & 79.15 & 1252 & 242 & 64.40 & 2582 & 324 & 27.37 & 4132 & 166 & 33.73 & 4127 & 209 \\
\textit{Discrete CoT} & $\leq 1000$ & 67.17 & 599 & 535 & 60.60 & 879 & 627 & 28.06 & 986 & 469 & 26.51 & 984 & 711 \\
\textit{Discrete CoT} & $\leq 500$ & 76.65 & 438 & 337 & 63.20 & 494 & 556 & 22.20 & 501 & 472 & 25.30 & 501 & 646 \\
\rowcolor{black!3}
\textit{MSE regression} & $\leq 32$ & \textbf{82.41} & 4.0 & 300 & 62.40 & 4.1 & 550 & 27.88 & 4.6 & 604 & \underline{37.35} & 4.6 & 669 \\
\rowcolor{black!3}
\textit{ATF} (Ours) & 0 & 28.05 & 0 & 698 & 30.00 & 0 & 733 & 14.63 & 0 & 747 & 15.66 & 0 & 740 \\
\rowcolor{black!3}
\textit{ATF} (Ours) & $\leq 32$ & 79.76 & 4.2 & 299 & \underline{66.80} & 4.9 & 549 & \underline{30.46} & 6.5 & 607 & \textbf{42.17} & 6.4 & 699 \\
\rowcolor{black!3}
\textit{ATF-RL} (Ours) & $\leq 32$ & \underline{81.73} & 4.0 & 289 & \textbf{69.00} & 4.9 & 493 & \textbf{32.19} & 6.1 & 631 & \textbf{42.17} & 5.7 & 654 \\
\midrule
\multicolumn{14}{l}{\textbf{\textit{Qwen3-4B}}} \\
\addlinespace[1pt]
\textit{Discrete CoT} & $\infty$ & 91.21 & 804 & 267 & 79.40 & 2020 & 417 & 44.58 & 3721 & 243 & 56.63 & 3469 & 397 \\
\textit{Discrete CoT} & $\leq 1000$ & 47.84 & 551 & 721 & 49.20 & 865 & 391 & 20.14 & 988 & 101 & 25.30 & 983 & 140 \\
\textit{Discrete CoT} & $\leq 500$ & 72.18 & 430 & 264 & 36.00 & 492 & 125 & 6.88 & 501 & 20 & 13.25 & 501 & 30 \\
\rowcolor{black!3}
\textit{MSE regression} & $\leq 32$ & 90.98 & 4.0 & 295 & 75.40 & 15.4 & 546 & 39.59 & 8.6 & 580 & 44.58 & 4.0 & 714 \\
\rowcolor{black!3}
\textit{ATF} (Ours) & 0 & 48.07 & 0 & 504 & 25.60 & 0 & 685 & 15.49 & 0 & 662 & 19.28 & 0 & 629 \\
\rowcolor{black!3}
\textit{ATF} (Ours) & $\leq 32$ & \underline{91.28} & 4.1 & 291 & \underline{78.40} & 6.8 & 533 & \underline{42.51} & 8.1 & 601 & \underline{56.63} & 8.0 & 728 \\
\rowcolor{black!3}
\textit{ATF-RL} (Ours) & $\leq 32$ & \textbf{92.04} & 5.3 & 283 & \textbf{78.60} & 12.9 & 493 & \textbf{44.41} & 21.4 & 622 & \textbf{57.83} & 20.8 & 666 \\
\bottomrule
\end{tabular}
\end{table*}

\begin{table}[t]
\centering
\hypersetup{hidelinks}
\small
\caption{\textbf{Multimodal next-thought prediction unlocks strong test-time scaling.}
ATF is competitive with prior methods at Pass@1, but the gap widens sharply as more latent trajectories are sampled.
At Pass@100, ATF outperforms every baseline, reaching $83.2$ on MATH-500, $97.6$ on GSM8K, and $52.4$ on OlympiadBench---a $34.6$ point gain over the strongest prior result on OlympiadBench.
The result suggests that ATF does not merely improve a single reasoning trajectory; \textbf{its multimodal next-thought distribution reaches diverse solution modes that become increasingly valuable under test-time sampling.}
With ATF (Greedy), the final answer is decoded deterministically, isolating continuous thinking as the source of this diversity.
All methods use LLaMA-3.1-8B; baseline scores are reported by LaDiR~\citep{kang2025ladir}.
Bold marks the best result for each metric and benchmark.}
\label{tab:rebuttal-ladir}
\definecolor{ourscol}{RGB}{230,242,255}
\setlength{\tabcolsep}{0pt}
\renewcommand{\arraystretch}{1.02}
\begin{tabular}{@{}>{\raggedright\arraybackslash}p{0.17\linewidth}>{\raggedright\arraybackslash}p{0.25\linewidth}*{6}{>{\centering\arraybackslash}p{0.096666\linewidth}}@{}}
\toprule
\multirow{2}{*}{\textbf{Family}} & \multirow{2}{*}{\textbf{Method}}
& \multicolumn{2}{c}{\textbf{MATH-500}} & \multicolumn{2}{c}{\textbf{GSM8K}} & \multicolumn{2}{c}{\textbf{OlympiadBench}} \\
\cmidrule(lr){3-4}\cmidrule(lr){5-6}\cmidrule(lr){7-8}
& & {\footnotesize\textbf{Pass@1}} & {\footnotesize\textbf{Pass@100}}
& {\footnotesize\textbf{Pass@1}} & {\footnotesize\textbf{Pass@100}}
& {\footnotesize\textbf{Pass@1}} & {\footnotesize\textbf{Pass@100}} \\
\midrule
\multirow{4}{*}{\textit{Hidden states}}
& \textit{iCoT}~{\scriptsize\citep{icot}} & $35.2$ & $37.9$ & $61.8$ & $63.9$ & $4.3$ & $7.1$ \\
& \textit{Coconut}~{\scriptsize\citep{hao2022training}} & $37.3$ & $39.3$ & $68.3$ & $74.3$ & $5.8$ & $6.3$ \\
& \textit{CODI}~{\scriptsize\citep{shen2025codi}} & $38.5$ & $45.1$ & $76.3$ & $81.7$ & $7.6$ & $14.8$ \\
& \textit{TaH+}~{\scriptsize\citep{tahplus}} & $46.1$ & $49.4$ & $\mathbf{85.9}$ & $89.7$ & $12.2$ & $15.2$ \\
\midrule
\multirow{2}{*}{\textit{Token-based}}
& \mbox{\textit{Disc. Latent}~{\scriptsize\citep{discrete_latent_thought}}} & $43.2$ & $47.3$ & $83.9$ & $88.6$ & $\mathbf{13.3}$ & $17.8$ \\
& \textit{Soft Think}~{\scriptsize\citep{softthinking}} & $44.3$ & $46.7$ & $83.7$ & $86.6$ & $10.4$ & $13.1$ \\
\midrule
\multirow{2}{*}{\textit{Block diffusion}}
& \textit{LaDiR w/o Stage 2} & $30.7$ & $35.8$ & $57.8$ & $62.6$ & $5.9$ & $10.5$ \\
& \textit{LaDiR}~{\scriptsize\citep{kang2025ladir}} & $\mathbf{46.2}$ & $63.7$ & $84.8$ & $93.7$ & $12.9$ & $15.3$ \\
\midrule
\rowcolor{black!3}
\cellcolor{white} & \textit{\textbf{ATF}} (Greedy) & $43.2$ & \cellcolor{ourscol}$71.6$ & $84.9$ & \cellcolor{ourscol}$93.1$ & $13.0$ & \cellcolor{ourscol}$38.9$ \\
\rowcolor{black!3}
\cellcolor{white}\multirow{-2}{*}{\textit{Next-thought flow}}
& \textit{\textbf{ATF}} & $42.2$ & \cellcolor{ourscol}$\mathbf{83.2}$ & $84.6$ & \cellcolor{ourscol}$\mathbf{97.6}$ & $12.7$ & \cellcolor{ourscol}$\mathbf{52.4}$ \\
\bottomrule
\end{tabular}
\end{table}

\paragraph{Evaluation metrics.} We report final-answer accuracy, mean thinking length $N_{\text{think}}$ (text tokens for CoT; latent steps for continuous methods), and mean answer length $N_{\text{ans}}$ in text tokens. Realized lengths differ from budgets and do not directly measure compute cost. Pass@1 and Pass@100 assess single-sample success and solution coverage with \model{}; Pass@$K$ measures the fraction of problems with at least one correct answer among $K$ samples. Trajectory diversity measures mean pairwise cosine distance. For Table~\ref{tab:main-results}, we rescore saved answers with a common Math-Verify v0.9.0 grader, retaining original generation settings; accuracy and lengths use the same outputs.

\paragraph{Implementation details.} We use \texttt{Qwen3-1.7B-Base} and \texttt{Qwen3-4B-Base}~\citep{qwen3}, filtering training examples whose thinking trace exceeds $4096$ tokens. CR-VAE uses chunk size $512$ and a compression curriculum from $2{:}1$ to $128{:}1$, ending with $4$ registers per chunk; it is trained for one epoch at learning rate $3\times10^{-5}$. We then freeze CR-VAE and train \model{} for one epoch on its latent trajectories. Unless otherwise stated, continuous methods use adaptive stopping with at most $32$ latent steps. Qwen3 ATF-RL rows use answer-policy refinement, separately from the joint latent/text RL model in the LLaMA-3.1-8B comparison.

\vspace{-3mm}
\subsection{Main Results}

\paragraph{\model{} improves the compact-reasoning frontier.} Table~\ref{tab:main-results} shows that \model{} improves over MSE on the harder 1.7B settings and all 4B settings while using roughly $4$--$8$ latent steps. 
Additional Pass@16 and self-consistency results are reported in Appendix~\ref{sec:multi-sample-analysis}.

\paragraph{Broader solution coverage through latent exploration.}
Table~\ref{tab:rebuttal-ladir} compares LLaMA-3.1-8B methods using baseline
scores reported in LaDiR~\citep{kang2025ladir}; ATF uses DART-MATH-derived data
and RL refinement. With $T_{\mathrm{text}}=0.7$, ATF achieves the highest
Pass@100 among the listed methods on all three benchmarks despite comparable
or lower Pass@1 than LaDiR. On MATH-500, ATF reaches $42.2\%$ Pass@1 versus
LaDiR's $46.2\%$, but $83.2\%$ Pass@100 versus $63.7\%$. On OlympiadBench,
nearly equal Pass@1 ($12.7\%$ versus $12.9\%$) accompanies substantially
higher Pass@100 ($52.4\%$ versus $15.3\%$). With greedy answer decoding,
ATF already exceeds LaDiR's Pass@100 on both benchmarks ($71.6\%$ and
$38.9\%$); GSM8K coverage is similar ($93.1\%$ versus $93.7\%$).
In this greedy-text setting, variation arises from latent initial noise,
showing that sampling latent trajectories alone can produce useful alternative
continuations. These results support useful latent exploration, complementing the matched
MSE comparison. Cross-paper protocol alignment and additional temperature
results appear in Appendix~\ref{sec:baseline-protocols} and
Table~\ref{tab:latent-baselines-full}.

\begin{table}[t]
\centering
\hypersetup{hidelinks}
\small
\caption{\textbf{Comparison with CoLaR.} CoLaR~\citep{tan2025think} and ATF use LLaMA-3.2-1B-Instruct under the GSM8K-Aug/MATH recipe; mean $\pm$ standard deviation over five runs. We directly reuse CoLaR's number and setting, which is the cloest related work that use a Gaussian distribution (single mode distribution) to model the next prediction. For the \model{} here we use exactly CoLaR's setting for fair comparison. Bold marks the better mean within each dataset.}
\label{tab:rebuttal-colar}
\setlength{\tabcolsep}{0pt}
\renewcommand{\arraystretch}{1.04}
\begin{tabular}{@{}>{\raggedright\arraybackslash}p{0.18\linewidth}>{\raggedright\arraybackslash}p{0.12\linewidth}*{4}{>{\centering\arraybackslash}p{0.175\linewidth}}@{}}
\toprule
\multirow{2}{*}{\textbf{Distribution}}
& \multirow{2}{*}{\textbf{Method}}
& \multicolumn{2}{c}{\textbf{GSM8K}} & \multicolumn{2}{c}{\textbf{MATH}} \\
\cmidrule(lr){3-4} \cmidrule(lr){5-6}
& & \textbf{Acc. (\%)} $\uparrow$ & \textbf{Latent steps} $\downarrow$
& \textbf{Acc. (\%)} $\uparrow$ & \textbf{Latent steps} $\downarrow$ \\
\midrule
\textit{Single Gaussian} & \textit{CoLaR} & $26.8\pm0.17$ & $\mathbf{5.57}\pm0.02$
& $7.08\pm0.07$ & $16.1\pm0.14$ \\
\rowcolor{black!3}
\cellcolor{white}\textit{Flow-based} & \textit{\textbf{ATF}} & $\mathbf{30.17}\pm1.26$ & $6.43\pm0.01$
& $\mathbf{8.59}\pm0.05$ & $\mathbf{7.17}\pm0.03$ \\
\bottomrule
\end{tabular}
\end{table}

\paragraph{Comparison with a Gaussian latent baseline.}
CoLaR~\citep{tan2025think} already models a stochastic next latent with a
Gaussian head, making it a stronger comparison than deterministic regression
alone. Under the shared LLaMA-3.2-1B-Instruct backbone, GSM8K-Aug/MATH
training data, and evaluation protocol in Table~\ref{tab:rebuttal-colar},
ATF improves accuracy by $3.37$ percentage points on GSM8K and $1.51$ on
MATH. On MATH, it also uses $7.17$ rather than $16.1$ latent steps on
average; on GSM8K, it uses slightly more steps ($6.43$ versus $5.57$).
This comparison extends the evidence for generative continuous reasoning
beyond the deterministic MSE baseline.

\Needspace{23\baselineskip}
\begin{wrapfigure}{r}{0.43\textwidth}
    \centering\small
    \setlength{\abovecaptionskip}{3pt}
    \setlength{\belowcaptionskip}{0pt}
    \includegraphics[width=\linewidth]{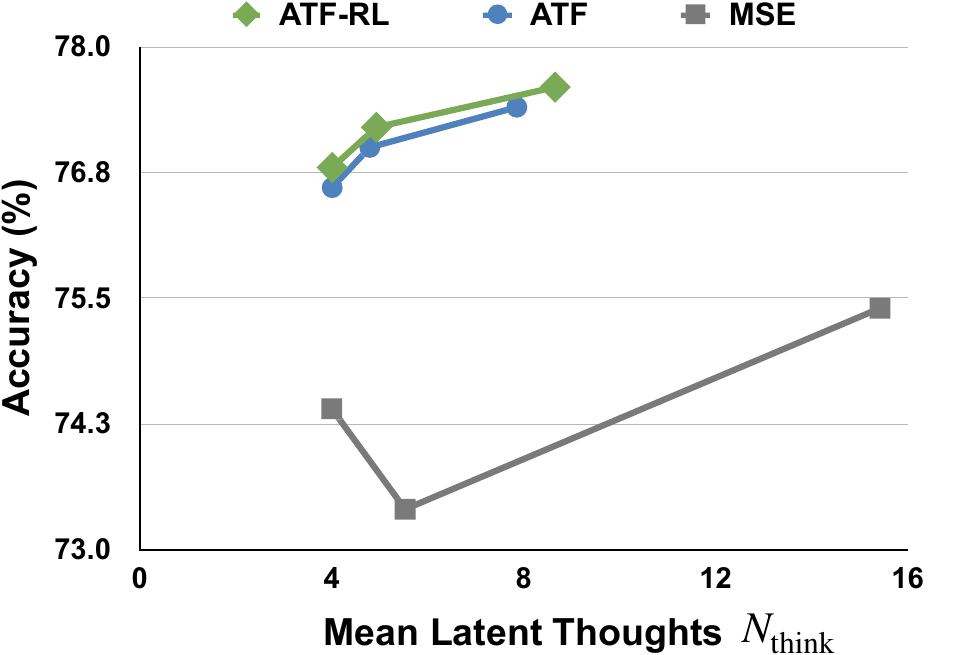}
    \vspace{-3mm}
    \caption{\textbf{\model{} benefits from more latent steps.}
    MATH-500 accuracy at 4B for stopping margins $m\in\{0,4,8\}$, with $\gamma=1$ and greedy answer decoding.
    The horizontal axis reports realized mean latent length. ATF achieves a stronger accuracy–latent-length trade-off than the matched regression baseline in this sweep.}
    \label{fig:latent-scaling}
\end{wrapfigure}

\paragraph{Additional continuous thoughts improve reasoning within a compact budget.} Figure~\ref{fig:latent-scaling} varies the stopping margin on MATH-500. For this sweep, stopping is checked every four latent steps: the stop token must be the top prediction with a top-1/top-2 logit gap above $m$, subject to a cap of $32$ steps. Over the displayed range, \model{} improves from $76.6\%$ to $77.4\%$ as its mean latent length increases from $4.01$ to $7.86$ steps; ATF-RL follows a similar trend. At $m=8$, MSE uses $15.42$ steps and reaches $75.4\%$, whereas \model{} attains higher accuracy with about half as many steps. This comparison supports the utility of the learned continuous thoughts beyond the default stopping point.

\paragraph{CFG controls exploration within the learned distribution.} Table~\ref{tab:cfg-tradeoff} examines this coverage--accuracy trade-off by varying CFG for Qwen3-4B-Base \model{} on AIME~\citep{maa_competitions}. Diversity uses normalized latent trajectories to measure directional spread rather than latent magnitude. As $\gamma$ increases, Pass@1 improves initially while Pass@16 and Diversity fall.

\WFclear
\begin{table}[H]
\centering\small
\caption{\textbf{CFG provides a built-in control over latent mode exploration.}
For the 4B \model{} on AIME, the guidance scale $\gamma$ trades off single-sample accuracy against sampled solution coverage.
Larger $\gamma$ concentrates generation around high-confidence continuations, favoring exploitation, whereas smaller $\gamma$ preserves more diverse latent modes for exploration.
Bold marks the best result for each metric.}
\label{tab:cfg-tradeoff}
\setlength{\tabcolsep}{3pt}
\renewcommand{\arraystretch}{1.08}
\begin{tabular*}{\linewidth}{@{\extracolsep{\fill}}c rrr rrr@{}}
\toprule
\multirow{2}{*}{\textbf{$\gamma$}}
& \multicolumn{3}{c}{\textbf{AIME 24}}
& \multicolumn{3}{c}{\textbf{AIME 25}} \\
\cmidrule(lr){2-4}\cmidrule(lr){5-7}
& \textbf{Pass@1} & \textbf{Pass@16} & \textbf{Diversity}
& \textbf{Pass@1} & \textbf{Pass@16} & \textbf{Diversity} \\
\midrule
1.0 & 8.54 & \textbf{30.00} & \textbf{0.7503} & 6.25 & \textbf{23.33} & \textbf{0.7518} \\
2.5 & 9.38 & 26.67 & 0.6357 & \textbf{7.08} & 20.00 & 0.6407 \\
5.0 & \textbf{9.58} & 26.67 & 0.4147 & 6.88 & 20.00 & 0.4499 \\
\bottomrule
\end{tabular*}
\end{table}

\newsavebox{\expvispanelbox}
\newcommand{\expvisrowtitle}[1]{%
    {\setlength{\fboxsep}{1.5pt}%
    \colorbox{black!6}{\parbox{\dimexpr\textwidth-2\fboxsep\relax}{%
        \centering\footnotesize\bfseries\strut #1}}}\par\vspace{0.5mm}%
}

\textbf{Generative next thoughts preserve multiple modes.} Figure~\ref{fig:mode-preservation-control}(a--c) illustrates how the learning objective shapes next-thought geometry: MSE regression concentrates near a single continuation, while \model{} retains multiple modes. Table~\ref{tab:head-decoded-text} gives the decoded continuations. These qualitative examples complement the sampled solution-coverage results.
\begin{table}[H]
    \centering
    \footnotesize
    \definecolor{headshared}{HTML}{246B8E}
    \definecolor{headdivergent}{HTML}{A05A2C}
    \newcommand{\headcommon}[1]{{\color{headshared}\bfseries\boldmath #1}}
    \newcommand{\headalternative}[1]{{\color{headdivergent}\bfseries\boldmath #1}}
    \setlength{\tabcolsep}{4pt}
    \renewcommand{\arraystretch}{1.10}
    \caption{\textbf{Decoded latent thoughts reveal regression collapse versus generative next-thought diversity.} MSE regression produces similar premise-restatement fragments, while \model{} realizes distinct continuations from the same autoregressive state. Blue highlights shared premise restatements; orange highlights contrasting continuations.}
    \label{tab:head-decoded-text}
    \begin{tabular}{@{}
        >{\raggedright\arraybackslash}p{0.17\linewidth}
        >{\raggedright\arraybackslash}p{0.065\linewidth}
        >{\raggedright\arraybackslash}p{\dimexpr0.765\linewidth-4\tabcolsep\relax}@{}}
    \toprule
    \textbf{Method} & \textbf{Trace} & \textbf{Decoded excerpt} \\
    \midrule
    MSE regression & A & ``Okay, so I need to find all natural numbers $N$ such that \headcommon{$N(N-101)$ is a perfect square} \ldots First, I remember that \headcommon{$N(N-101)$ needs to be a square} \ldots'' \\
    MSE regression & B & ``If $N$ is a natural number, then \headcommon{$N(N-101)$ must equal the square} \ldots So, \headcommon{$N(N-101)$ must be a perfect square} \ldots'' \\
    \midrule
    \model{} & A & ``\headalternative{Expressed in the form $N(N-101)=x^2$} \ldots $N=x^2$ is a perfect square \ldots'' \\
    \model{} & B & ``\headalternative{$N$ is the perimeter of a square} \ldots identify which numbers from $1\times 1$ to $69$ are squares.'' \\
    \bottomrule
    \end{tabular}
\end{table}

\begingroup
\setlength{\intextsep}{1pt}
\begin{figure}[!t]
    \centering
    \setlength{\parskip}{0pt}
    \expvisrowtitle{Multi-mode seeking: broader coverage than MSE regression}
    \sbox{\expvispanelbox}{\includegraphics[height=0.226\textwidth]{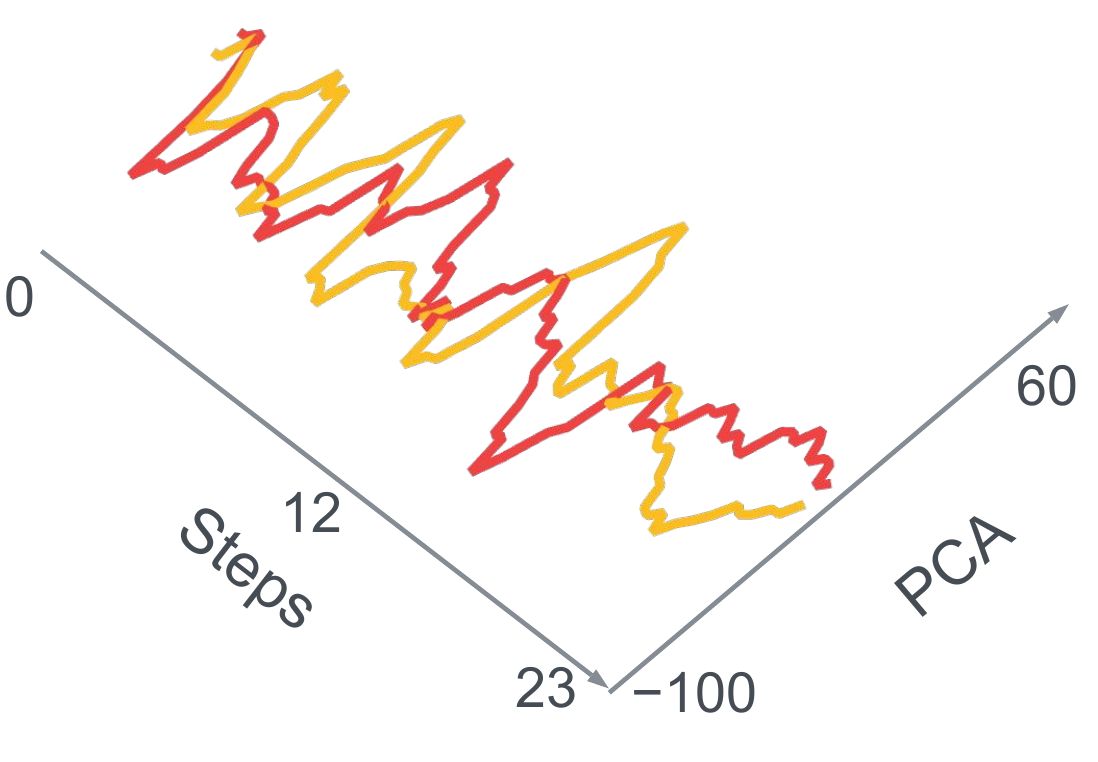}}%
    \begin{minipage}[t]{\wd\expvispanelbox}
        \centering
        \usebox{\expvispanelbox}\\[0.2mm]
        {\footnotesize (a) Ground-truth CoT}
    \end{minipage}%
\hfill
    \sbox{\expvispanelbox}{\includegraphics[height=0.226\textwidth]{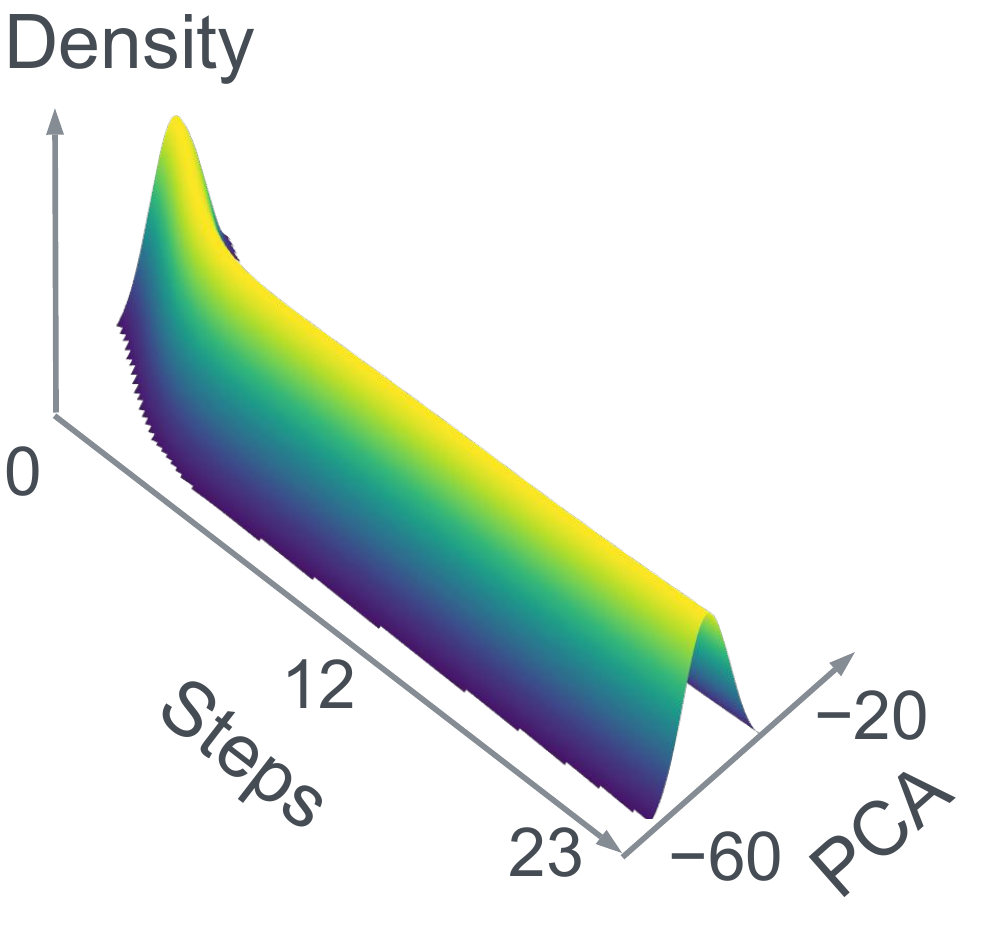}}%
    \begin{minipage}[t]{\wd\expvispanelbox}
        \centering
        \usebox{\expvispanelbox}\\[0.2mm]
        {\footnotesize (b) MSE regression}
    \end{minipage}%
\hfill
    \sbox{\expvispanelbox}{\includegraphics[height=0.226\textwidth]{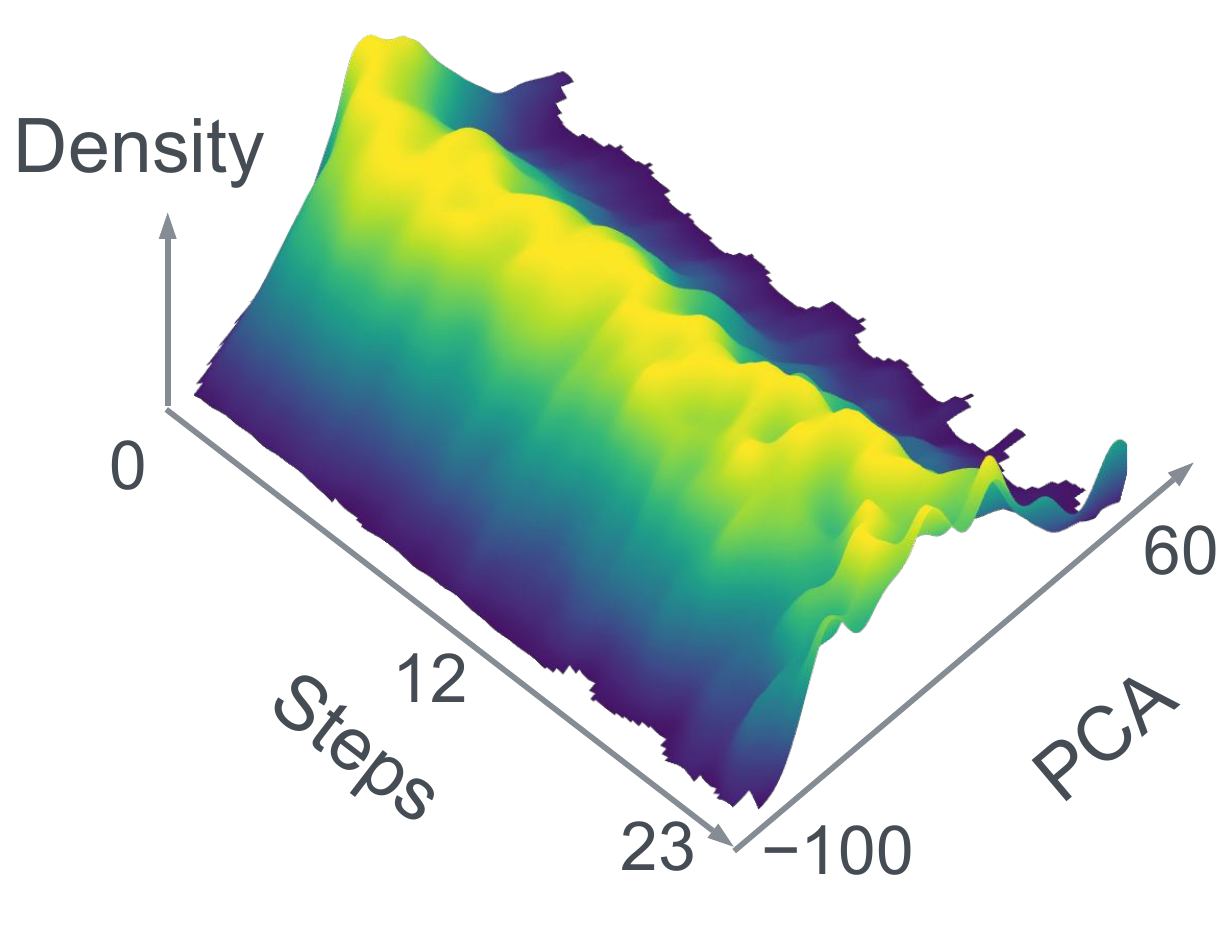}}%
    \begin{minipage}[t]{\wd\expvispanelbox}
        \centering
        \usebox{\expvispanelbox}\\[0.2mm]
        {\footnotesize (c) \model{}}
    \end{minipage}%
\par\vspace{0.8mm}
    \expvisrowtitle{Controllable mode seeking: exploration--exploitation via CFG}
    \sbox{\expvispanelbox}{\includegraphics[height=0.252\textwidth]{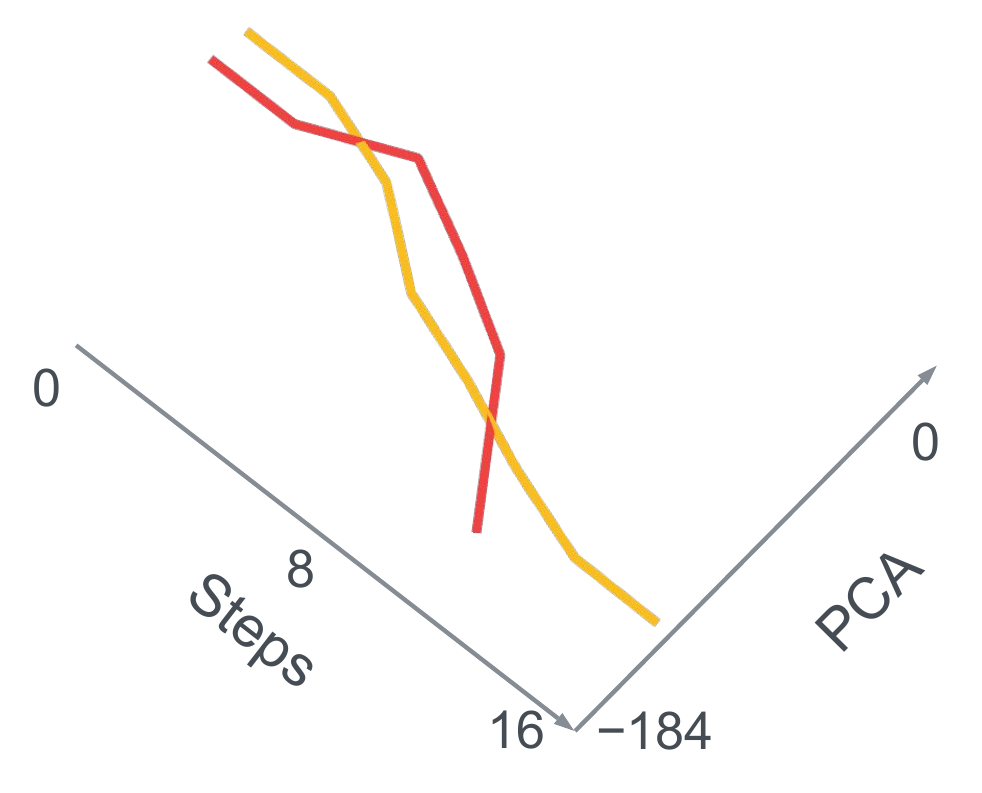}}%
    \begin{minipage}[t]{\wd\expvispanelbox}
        \centering
        \usebox{\expvispanelbox}\\[0.2mm]
        {\footnotesize (d) Ground-truth CoT}
    \end{minipage}%
\hfill
    \sbox{\expvispanelbox}{\includegraphics[height=0.252\textwidth]{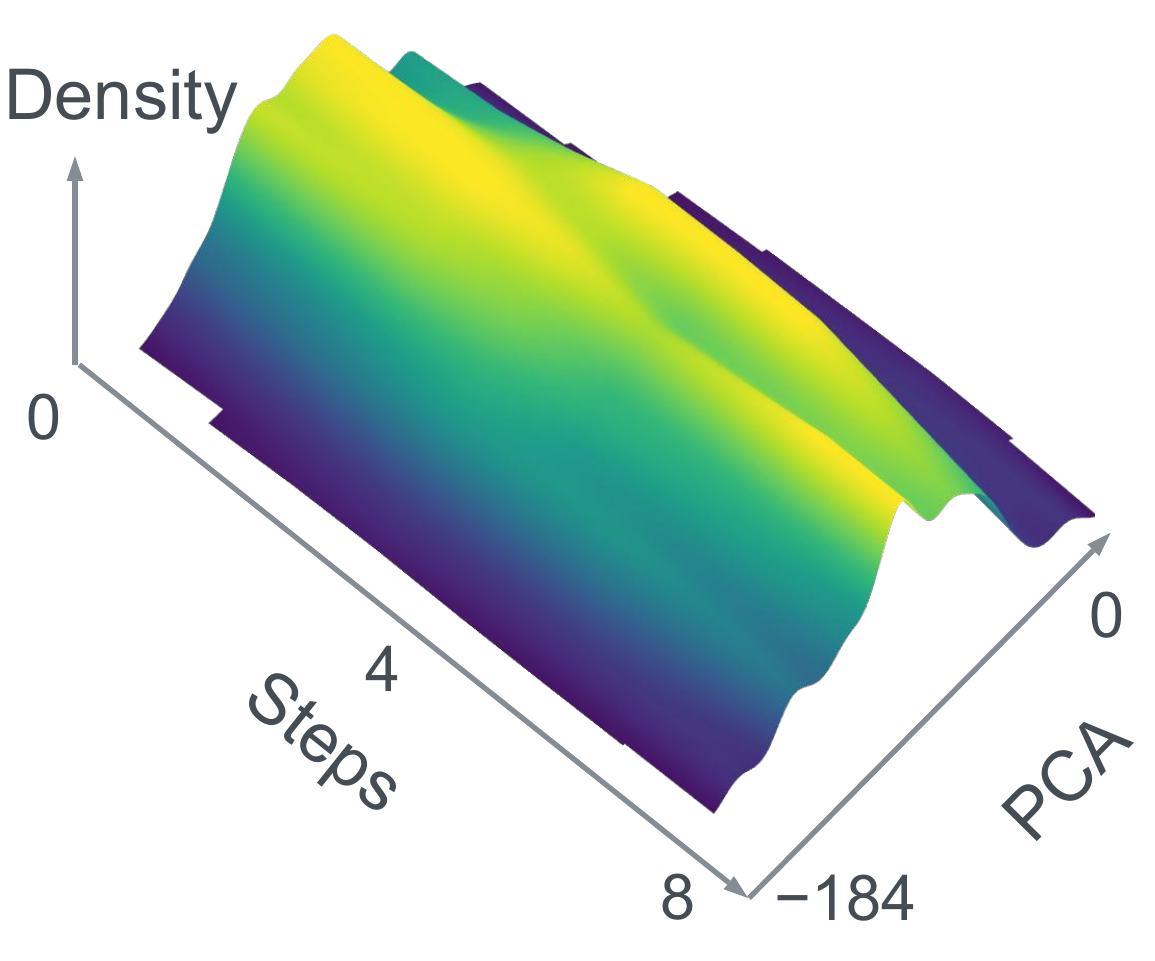}}%
    \begin{minipage}[t]{\wd\expvispanelbox}
        \centering
        \usebox{\expvispanelbox}\\[0.2mm]
        {\footnotesize (e) $\gamma=1$, Pass@1 = 25\%}
    \end{minipage}%
\hfill
    \sbox{\expvispanelbox}{\includegraphics[height=0.252\textwidth]{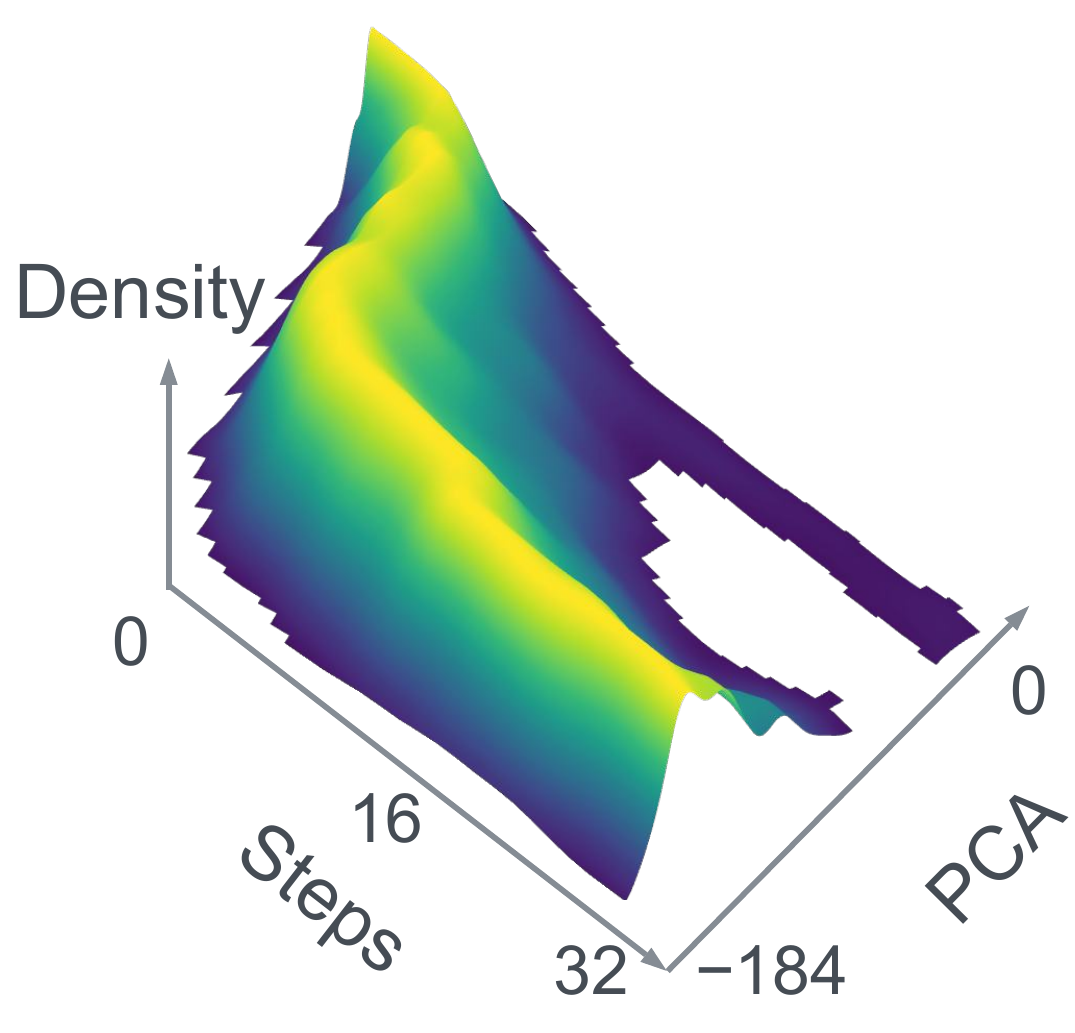}}%
    \begin{minipage}[t]{\wd\expvispanelbox}
        \centering
        \usebox{\expvispanelbox}\\[0.2mm]
        {\footnotesize (f) $\gamma=5$, Pass@1 = 69\%}
    \end{minipage}
    \caption{PCA feature visualizations of \textbf{\model{}'s latent reasoning modes.} We sample 500 times to estimate the distribution for our diffusion model, and plot their density map. Compared with MSE regression, \model{} preserves more diverse next-thought continuations (top). Our diffusion head design allows us to apply the classifier-free guidance then adjusts the concentration of this learned distribution, trading broader exploration for more focused exploitation (bottom). The rows illustrate different questions; Pass@1 refers to the bottom-row example. Tables~\ref{tab:head-decoded-text} and~\ref{tab:cfg-decoded-text} give decoded excerpts.}
    \label{fig:mode-preservation-control}
\end{figure}
\paragraph{Guidance concentrates decoded continuations.} Figure~\ref{fig:mode-preservation-control}(d--f) illustrates the shift toward exploitation. In Table~\ref{tab:cfg-decoded-text}, low-guidance samples follow different row constructions, while both high-guidance samples reproduce the same initial row sequence.

\begin{table}[H]
    \centering
    \footnotesize
    \definecolor{cfgdivergent}{HTML}{A05A2C}
    \definecolor{cfgshared}{HTML}{246B8E}
    \newcommand{\cfgalternative}[1]{{\color{cfgdivergent}\bfseries\boldmath #1}}
    \newcommand{\cfgcommon}[1]{{\color{cfgshared}\bfseries\boldmath #1}}
    \setlength{\tabcolsep}{4pt}
    \renewcommand{\arraystretch}{1.15}
    \caption{\textbf{Stronger CFG concentrates decoded continuations around the reference reasoning pattern.} Low guidance produces more varied, sometimes off-mode continuations, while high guidance favors similar row-pattern reconstructions. Orange marks contrasting low-guidance fragments; blue marks the repeated high-guidance row sequence.}
    \label{tab:cfg-decoded-text}
    \begin{tabular}{@{}
        >{\raggedright\arraybackslash}p{0.07\linewidth}
        >{\raggedright\arraybackslash}p{0.055\linewidth}
        >{\raggedright\arraybackslash}p{\dimexpr0.875\linewidth-4\tabcolsep\relax}@{}}
    \toprule
    \textbf{CFG} $\gamma$ & \textbf{Trace} & \textbf{Decoded excerpt} \\
    \midrule
    $1$ & A & ``Okay, so the problem is about finding the units digit \ldots If $n=7$, then \mbox{\cfgalternative{row 7 is $(1,5,6,5,1)$}} \ldots'' \\
    \addlinespace[2pt]
    $1$ & B & ``Okay, let's see \ldots triangular array with 15th row \ldots additional entries are constructed by \mbox{\cfgalternative{reflecting a row with TicTacToe}} \ldots'' \\
    \midrule
    $5$ & A & ``Alright, let's try to figure out this problem \ldots The first rows are
    \cfgcommon{$1$, $1\ 1$, $1\ 3\ 1$, $1\ 5\ 5\ 1$, $1\ 7\ 11\ 7\ 1$} \ldots'' \\
    \addlinespace[2pt]
    $5$ & B & ``Alright, let's try to figure out this problem \ldots The rows are
    \cfgcommon{$1$, $1\ 1$, $1\ 3\ 1$, $1\ 5\ 5\ 1$, $1\ 7\ 11\ 7\ 1$} \ldots'' \\
    \bottomrule
    \end{tabular}
\end{table}

\endgroup

\vspace{-3mm}
\paragraph{Inference cost of continuous thoughts.}
In a separate Qwen3-4B profiling study on one RTX PRO 6000 (batch size 1), ATF uses one backbone step and 24 lightweight flow-head steps per thought. Length-based estimates reduce MATH-500 end-to-end time from $31.89$\,s for text CoT to $7.19$\,s for ATF ($4.44\times$). Direct timing measurements agreed within $3.5\%$ of the estimates. Appendix~\ref{sec:inference-efficiency} reports all four benchmarks and a matched per-thought compute comparison.
\vspace{-3mm}

\section{Conclusion}
\vspace{-3mm}
We introduced \model{}, which treats the next continuous thought as a multimodal distribution rather than a single prediction.
By combining a causal reasoning backbone with a lightweight generative head, \model{} preserves autoregressive reasoning while allowing multiple plausible continuations to remain possible.
Across challenging reasoning tasks, this design improves accuracy, supports test-time scaling, and yields substantially greater solution coverage through diverse latent trajectories.
Our results suggest that modeling uncertainty over what to think next is a key ingredient for effective continuous reasoning.

\subsection*{AI use statement}
Generative AI assistance was used to aid and polish writing, revise the
manuscript's text, organization, and presentation, and support literature
retrieval and discovery. LLM assistance was also used for code generation
and experiment execution.
It was also used to draft the reinforcement-learning method description
from the authors' slides and implementation, adapt the manuscript's LaTeX
submission format, and transcribe and format previously reported experimental
results as tables. The scientific ideas, experimental design, and interpretation
of the results originated with the authors.

\appendix
\section{Implementation Details for RL Refinement}
\label{sec:rl-implementation}
The optional RL extension in Section~\ref{sec:atf-rl} refines the existing
multimodal next-thought model using outcome rewards. This appendix specifies
the transition scores and optimization details behind the compact main-text
objective. The CR-VAE remains fixed, and $\theta$ collects the trainable
backbone and flow-head parameters.

For $N,M>1$, the centered advantages in the main text are standardized by the
sample standard deviation across latent-mean rewards and within-latent answer
rewards, respectively, with a small stabilizer and symmetric clipping. When
$N=1$, only the within-trajectory text advantage is used. A flat $M=1$ variant
uses the ordinary group-relative advantage across complete rollouts for both
terms.

\paragraph{Scoring stochastic flow actions.}
A flow-generated thought has no categorical next-token probability. Instead of
evaluating its intractable marginal density, we record stochastic denoising
transitions within each autoregressive thought. Let $u_{n,t,k}$ denote the
intermediate flow state for thought $t$ and denoising step $k$. At selected steps
$k\in\mathcal W$, noise injection gives a Gaussian transition
\begin{equation}
\begin{aligned}
    u_{n,t,k+1}&=\mu_\theta(u_{n,t,k},\tau_k,h_{n,t})+\sigma_k\xi_{n,t,k},
       &\xi_{n,t,k}&\sim\mathcal N(0,I),\\
    q_\theta(u_{n,t,k+1}\mid u_{n,t,k},h_{n,t})
       &=\mathcal N(\mu_\theta(u_{n,t,k},\tau_k,h_{n,t}),\sigma_k^2I).
\end{aligned}
\label{eq:rl-flow-transition}
\end{equation}
The mean is obtained from the flow vector field using the chosen stochastic
transition rule; $\sigma_k$ is fixed by the sampler. Remaining denoising steps
use deterministic integration. Once a thought is completed, it is appended to
the causal history before generating the next thought. During optimization,
recorded states are held fixed and rescored using the current backbone condition
and flow head; we do not backpropagate through the rollout sampler.

For latent dimension $d$, a dimension-averaged Gaussian log-density, up to terms
independent of the policy parameters, is
\begin{equation}
    \ell^{\mathrm{Gauss}}_{n,t,k}(\theta)
      =-\frac{\|u_{n,t,k+1}-\mu_\theta(u_{n,t,k},\tau_k,h_{n,t})\|_2^2}
              {2d\sigma_k^2}.
\end{equation}
Our implementation supports this variance-scaled score and a raw squared-error
score $s_{n,t,k}=-\|u_{n,t,k+1}-\mu_\theta\|_2^2/d$.
The latter is a policy-score surrogate: omitting $1/(2\sigma_k^2)$ changes the
relative weighting of denoising steps, so it is not an exact log-density.
Neither score is the marginal log-probability of the completed thought.

\paragraph{Policy update.}
For a stored action with current score $s$ and fixed behavior-policy score
$s_{\mathrm{old}}$, define
\begin{equation}
\begin{aligned}
    \rho&=\exp(s-s_{\mathrm{old}}),\\
    \mathcal C(\rho,A)
       &=\min\!\left(\rho A,
          \operatorname{clip}(\rho,1-\varepsilon,1+\varepsilon)A\right).
\end{aligned}
\end{equation}
For text actions, $s$ is the answer-token log-probability at the training
temperature. For latent actions it is the selected transition score above;
with raw squared error, $\rho$ is a surrogate score ratio. With the Gaussian
score, dimension averaging makes it a dimension-normalized likelihood ratio.
We maximize
\begin{equation}
    \mathcal J_{\mathrm{RL}}
      =\lambda_z\!\left\langle\mathcal C(\rho^z,A^{\mathrm{lat}})\right\rangle_z
       +\lambda_y\!\left\langle\mathcal C(\rho^y,A^{\mathrm{text}})\right\rangle_y,
\label{eq:rl-objective}
\end{equation}
where the two averages mask padding and normalize over valid stochastic
transitions and answer tokens, respectively. Separate clipping thresholds may
be used for the two score scales. Without clipping, the corresponding update is
the advantage-weighted score-function objective. The optional stop loss is an
advantage-weighted binary surrogate; stopping in the rollout remains a
deterministic margin decision and is not assigned an importance ratio.

\section{Baseline Definitions and Comparison Protocols}
\label{sec:baseline-protocols}

\paragraph{Matched regression baseline.}\label{sec:matched-regression}
To isolate the effect of generative next-thought modeling, we compare against a baseline with the same backbone, CR-VAE targets, hybrid sequence format, and stopping rule. The only change is the latent prediction head: instead of a flow model, the baseline predicts a deterministic vector $\hat{z}_t = g_\omega(h_t)$ and minimizes
\begin{equation}
    \mathcal{L}_{\mathrm{MSE}} =
    \mathbb{E}_{t}\left[\left\|g_\omega(h_t)-z_t\right\|_2^2\right].
\end{equation}
At inference, $\hat{z}_t$ is fed back as the next continuous thought. This baseline is therefore a point-estimate version of \model{} rather than a separately engineered system.

\paragraph{LLaMA-3.1-8B comparison.}
The non-ATF entries in Table~\ref{tab:rebuttal-ladir} are reported in
Table~1 of LaDiR~\citep{kang2025ladir}, including both LaDiR training stages.
MATH-500 denotes the 500-question subset reported as MATH in LaDiR.
Baselines retain their reported training procedures; this is a cross-paper
comparison, not a matched ablation of the prediction head or RL objective.

Table~\ref{tab:latent-baselines-full} reports both Pass@1 and Pass@100 for the
methods in the main-text comparison, including an additional text-temperature setting. ATF has the highest Pass@100 in this
roster, while the best Pass@1 scores belong to LaDiR on MATH-500, TaH+ on
GSM8K, and Discrete Latent on OlympiadBench.

\begin{table}[!htbp]
\centering
\small
\caption{\textbf{Complete single-sample and multi-sample comparison.} All methods use the LLaMA-3.1-8B backbone; ATF is trained on DART-MATH-derived data and RL-refined. Each cell reports Pass@1 / Pass@100 (\%). Baseline scores are reported by LaDiR. Bold marks the best value for each metric.}
\label{tab:latent-baselines-full}
\setlength{\tabcolsep}{6pt}
\renewcommand{\arraystretch}{1.08}
\begin{tabular}{@{}lccc@{}}
\toprule
Method & MATH-500 & GSM8K & OlympiadBench \\
\midrule
iCoT & $35.2\,/\,37.9$ & $61.8\,/\,63.9$ & $4.3\,/\,7.1$ \\
Coconut & $37.3\,/\,39.3$ & $68.3\,/\,74.3$ & $5.8\,/\,6.3$ \\
CODI & $38.5\,/\,45.1$ & $76.3\,/\,81.7$ & $7.6\,/\,14.8$ \\
Discrete Latent & $43.2\,/\,47.3$ & $83.9\,/\,88.6$ & $\mathbf{13.3}\,/\,17.8$ \\
Soft Think & $44.3\,/\,46.7$ & $83.7\,/\,86.6$ & $10.4\,/\,13.1$ \\
TaH+ & $46.1\,/\,49.4$ & $\mathbf{85.9}\,/\,89.7$ & $12.2\,/\,15.2$ \\
\midrule
LaDiR w/o Stage 2 & $30.7\,/\,35.8$ & $57.8\,/\,62.6$ & $5.9\,/\,10.5$ \\
LaDiR & $\mathbf{46.2}\,/\,63.7$ & $84.8\,/\,93.7$ & $12.9\,/\,15.3$ \\
\midrule
ATF ($T_{\mathrm{text}}=0$) & $43.2\,/\,71.6$ & $84.9\,/\,93.1$ & $13.0\,/\,38.9$ \\
ATF ($T_{\mathrm{text}}=0.7$) & $42.2\,/\,83.2$ & $84.6\,/\,97.6$ & $12.7\,/\,52.4$ \\
ATF ($T_{\mathrm{text}}=1$) & $41.4\,/\,\mathbf{84.8}$ & $84.5\,/\,\mathbf{98.3}$ & $12.6\,/\,\mathbf{52.5}$ \\
\bottomrule
\end{tabular}
\end{table}

ATF uses the LLaMA-3.1-8B backbone, DART-MATH-derived training data,
and joint latent/text RL. The main-table greedy and text-temperature-$0.7$
settings use 100 independently sampled latent trajectories per question,
with one answer per trajectory. Evaluation uses ODE latent sampling and DART
grading. OlympiadBench uses the 674-question split with multi-answer cases
retained. The temperature-$1$ row provides an additional evaluation setting.
LaDiR reports answer sampling temperature $0.7$.

\paragraph{LLaMA-3.2-1B-Instruct comparison.}
Table~\ref{tab:rebuttal-colar} compares ATF and CoLaR using a shared
LLaMA-3.2-1B-Instruct backbone, GSM8K-Aug/MATH training data, and evaluation
protocol. We report mean and standard deviation over five runs. The benchmark name MATH follows that experiment's reporting and is
not equated with the MATH-500 split above. Scores from the two tables are
therefore compared only within their respective experimental settings.

\section{CR-VAE Architecture}
\label{sec:crvae-architecture}

\begin{figure}[!htbp]
    \centering
    \includegraphics[width=\linewidth]{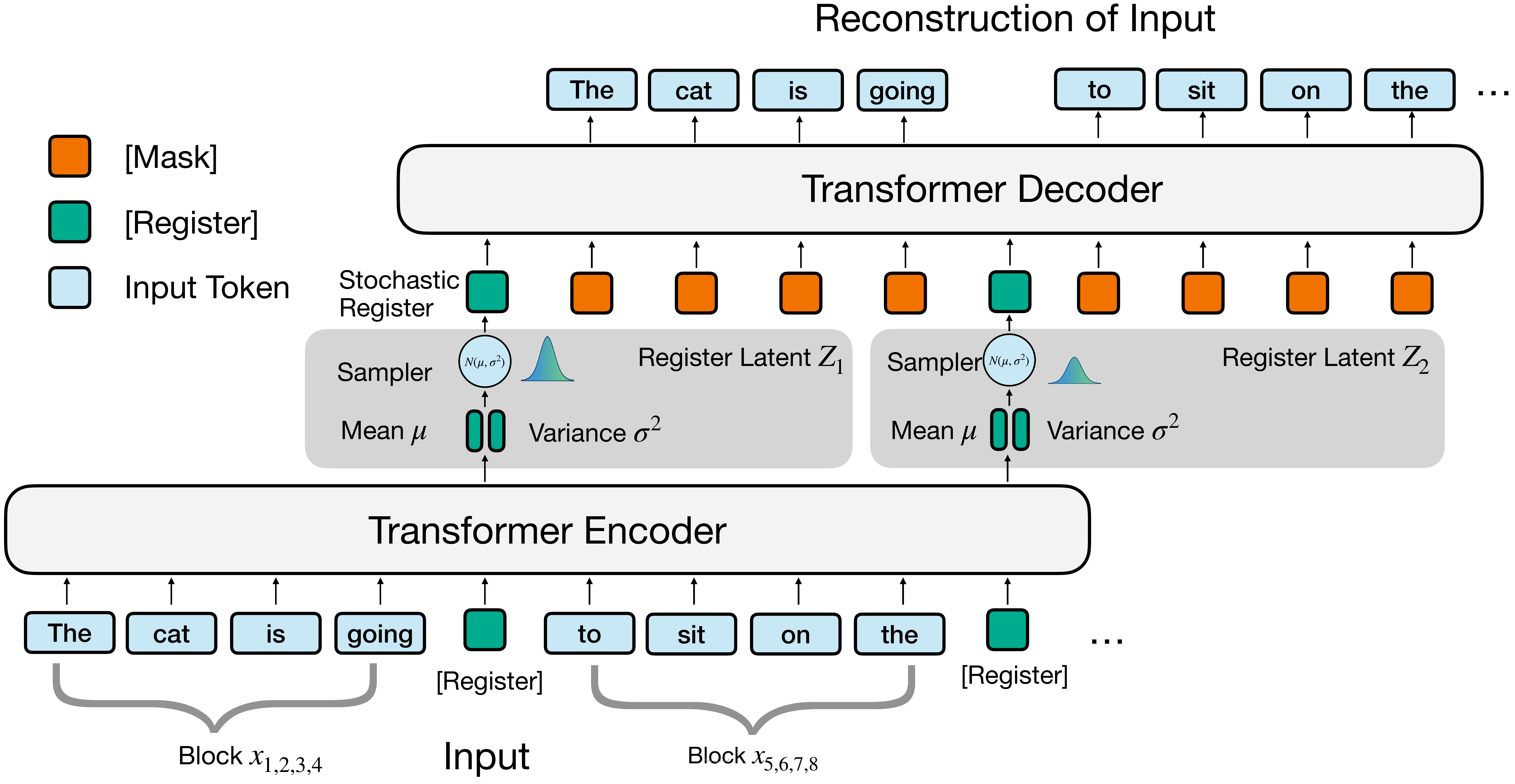}
    \caption{\textbf{Causal-Register VAE architecture.} Register tokens inserted after reasoning blocks provide the encoder states used to parameterize Gaussian posteriors. Sampled latent registers condition the transformer decoder, which reconstructs the input text using mask placeholders. Causal encoding preserves the ordering of the reasoning prefix; the resulting latent registers serve as continuous thought targets for ATF.}
    \label{fig:crvae-architecture}
\end{figure}

Figure~\ref{fig:crvae-architecture} shows the CR-VAE used to obtain continuous thought targets. As described in Section~\ref{sec:crvae}, causal registers summarize the reasoning prefix into stochastic latent variables. After reconstruction training, the encoder is frozen and used offline to produce the latent trajectories for ATF supervision; the decoder supports qualitative inspection of generated thoughts.

\section{Ablation Studies}
\label{sec:ablation-studies}

\subsection{CR-VAE Compression}

\paragraph{How compact should the latent thought space be?} On Qwen3-1.7B-Base, we ablate the CR-VAE compression ratio to test whether \model{} requires token-faithful reconstruction, or only a compact latent space that preserves the reasoning information needed for downstream answers. Table~\ref{tab:compression-ablation} shows a clear separation between reconstruction fidelity and reasoning utility: stronger compression substantially lowers autoregressive token reconstruction, yet downstream accuracy remains stable or improves. This suggests that discrete chain-of-thought contains substantial token-level redundancy for reasoning; \model{} can therefore reason efficiently in a compact latent space, avoiding the accumulation of errors that arises when long reasoning traces must be reproduced token by token.

\begin{table}[!htbp]
\centering
\footnotesize
\caption{\textbf{More compact CR-VAE latents can improve downstream reasoning despite lower token reconstruction at 1.7B.} Using the Qwen3-1.7B-Base backbone, AR reconstruction accuracy measures in-distribution reconstruction of discrete thoughts, while GSM8K and OlympiadBench report CFG = 1 reasoning accuracy. Higher VAE compression discards token-level surface form, yet downstream accuracy remains strong, indicating that compact latents preserve the reasoning structure needed by \model{}.}
\label{tab:compression-ablation}
\setlength{\aboverulesep}{0.35ex}
\setlength{\belowrulesep}{0.55ex}
\setlength{\tabcolsep}{4.0pt}
\renewcommand{\arraystretch}{1.12}
\begin{tabular}{c c c c}
\toprule
\textbf{VAE Comp. Ratio} & \textbf{AR Recon Acc.} & \textbf{GSM8K} & \textbf{Olympiad} \\
\specialrule{0.045em}{0.35ex}{0.6ex}
\textrm{16:1}  & \textbf{99.1} & 76.80 & 20.31 \\
\textrm{64:1}  & 75.2 & 78.09 & \textbf{22.03} \\
\textrm{128:1} & 25.7 & \textbf{80.52} & 21.17 \\
\bottomrule
\end{tabular}%
\end{table}

\section{Inference Efficiency}
\label{sec:inference-efficiency}

\paragraph{End-to-end cost.}
We profile reasoning and answer decoding separately on one NVIDIA RTX PRO 6000 at batch size 1. Table~\ref{tab:rebuttal-latency} combines profiled operator costs with actual generated lengths under each method's evaluation configuration; direct timing measurements agreed within $3.5\%$ of these estimates. This is a separate profiling study, rather than a remeasurement of the evaluation outputs selected for Table~\ref{tab:main-results}. ATF uses 24 flow-head denoising steps per latent thought. Answer decoding accounts for most of its estimated runtime, so shorter latent traces do not by themselves determine end-to-end speed.

\begin{table}[H]
\centering
\small
\caption{\textbf{End-to-end latency estimates for Qwen3-4B.} Reasoning, answer, and total times on one RTX PRO 6000, batch size 1. ATF uses 24 flow-head denoising steps per thought.}
\label{tab:rebuttal-latency}
\setlength{\tabcolsep}{4pt}
\begin{tabular*}{\linewidth}{@{\extracolsep{\fill}}llrrrr@{}}
\toprule
\textbf{Benchmark} & \textbf{Method} & \textbf{Think (s)} & \textbf{Answer (s)} & \textbf{Total (s)} & \textbf{Speedup} \\
\midrule
GSM8K & Text CoT & 10.53 & 3.49 & 14.02 & $1.00\times$ \\
& ATF & 0.22 & 3.80 & 4.02 & $3.49\times$ \\
\midrule
MATH-500 & Text CoT & 26.43 & 5.46 & 31.89 & $1.00\times$ \\
& ATF & 0.22 & 6.97 & 7.19 & $4.44\times$ \\
\midrule
OlympiadBench & Text CoT & 48.69 & 3.18 & 51.86 & $1.00\times$ \\
& ATF & 0.26 & 7.89 & 8.15 & $6.40\times$ \\
\midrule
AMC 12 & Text CoT & 45.60 & 5.21 & 50.81 & $1.00\times$ \\
& ATF & 0.41 & 9.56 & 9.97 & $5.09\times$ \\
\bottomrule
\end{tabular*}
\end{table}

\paragraph{Cost of generating one thought.}
Table~\ref{tab:rebuttal-per-thought-compute} isolates per-thought generation under the same LLaMA-3.1-8B backbone, $5.5{:}1$ compression, hardware, and batch size 1. ATF runs the backbone once per thought and performs its 24 denoising steps in the flow head, whereas LaDiR runs the backbone at each denoising step. This comparison measures the cost of the next-thought interface, independently of the number of thoughts or answer tokens generated for a benchmark.

\begin{table}[H]
\centering
\small
\caption{\textbf{Per-thought compute under matched conditions.} LLaMA-3.1-8B backbone, $5.5{:}1$ compression, identical hardware, and batch size 1. The step counts denote flow-head evaluations for ATF and full-backbone evaluations for LaDiR.}
\label{tab:rebuttal-per-thought-compute}
\begin{tabular}{@{}lrr@{}}
\toprule
\textbf{Method} & \textbf{GFLOPs / thought} & \textbf{Latency / thought (ms)} \\
\midrule
ATF & \textbf{25.2} & \textbf{35.8} \\
LaDiR, 10 steps & 139.6 & 41.4 \\
LaDiR, 50 steps & 697.9 & 207.0 \\
\bottomrule
\end{tabular}
\end{table}

\Needspace{20\baselineskip}
\section{Additional Sampling Analysis}
\label{sec:multi-sample-analysis}

\paragraph{\model{} provides broader sampled answer coverage than MSE regression.} On both MATH-500 and AMC 12, Table~\ref{tab:multi-sample} shows higher Pass@16 for \model{}. On MATH-500, \model{} also improves self-consistency accuracy ($73.00$ versus $69.00$), showing that broader exploration can increase both solution coverage and majority-vote success. On AMC 12, MSE has higher self-consistency despite lower Pass@16, illustrating that answer concentration and solution coverage need not improve together. The consistent coverage gain supports the usefulness of sampled latent alternatives. Figure~\ref{fig:mode-preservation-control}(a--c) and Table~\ref{tab:head-decoded-text} provide complementary qualitative illustrations of different continuations.

\begin{table}[H]
\centering
\footnotesize
\caption{\textbf{Multi-sample rollouts show broader answer coverage for \model{} at 4B.} Using the Qwen3-4B-Base backbone, we evaluate $K=16$ adaptive-stopping rollouts at temperature $0.1$ on \textbf{MATH-500} and \textbf{AMC 12}. Pass@16 measures whether any sampled answer is correct, while self-consistency accuracy uses majority voting across the same samples.}
\label{tab:multi-sample}
\setlength{\aboverulesep}{0.35ex}
\setlength{\belowrulesep}{0.55ex}
\setlength{\tabcolsep}{5.0pt}
\renewcommand{\arraystretch}{1.12}
\begin{tabular}{l cc cc}
\toprule
\textbf{Method}
& \multicolumn{2}{c}{\textbf{MATH-500}}
& \multicolumn{2}{c}{\textbf{AMC 12}} \\
\cmidrule(lr){2-3} \cmidrule(lr){4-5}
& \textbf{Pass@16} & \makecell{\textbf{Self-consistency}\\\textbf{Acc.}}
& \textbf{Pass@16} & \makecell{\textbf{Self-consistency}\\\textbf{Acc.}} \\
\specialrule{0.045em}{0.35ex}{0.6ex}
\textit{MSE regression} & 77.60 & 69.00 & 72.29 & \textbf{62.65} \\
\textit{\model{}} \textbf{(Ours)} & \textbf{79.80} & \textbf{73.00} & \textbf{74.70} & 57.83 \\
\bottomrule
\end{tabular}%
\end{table}

\end{document}